\PassOptionsToPackage{unicode}{hyperref}
\PassOptionsToPackage{hyphens}{url}
\documentclass[
]{article}
\usepackage{lmodern}
\usepackage{amssymb,amsmath}
\usepackage{ifxetex,ifluatex}
\ifnum 0\ifxetex 1\fi\ifluatex 1\fi=0 
  \usepackage[T1]{fontenc}
  \usepackage[utf8]{inputenc}
  \usepackage{textcomp} 
\else 
  \usepackage{unicode-math}
  \defaultfontfeatures{Scale=MatchLowercase}
  \defaultfontfeatures[\rmfamily]{Ligatures=TeX,Scale=1}
\fi
\IfFileExists{upquote.sty}{\usepackage{upquote}}{}
\IfFileExists{microtype.sty}{
  \usepackage[]{microtype}
  \UseMicrotypeSet[protrusion]{basicmath} 
}{}
\makeatletter
\@ifundefined{KOMAClassName}{
  \IfFileExists{parskip.sty}{%
    \usepackage{parskip}
  }{
    \setlength{\parindent}{0pt}
    \setlength{\parskip}{6pt plus 2pt minus 1pt}}
}{
  \KOMAoptions{parskip=half}}
\makeatother
\usepackage{xcolor}
\usepackage{fancyhdr} 

\fancypagestyle{firstpage}{%
  \fancyhf{}
  \fancyfoot[C]{\parbox{\textwidth}{\centering \footnotesize \textbf{Preprint version.} Published as: Giarmatzis, G., Horst, F., Slijepcevic, D., Aggeloussis, N., Papakostas, G. (2027). Retrieval-Augmented Generation for Clinical Decision Support in Gait Analysis: A Proof-of-Concept. In: Saraswat, M., Tiwari, R., Pavone, M.F., Patel, S.J. (eds) Proceedings of International Conference on Computational Intelligence. ICCI 2025. Lecture Notes in Networks and Systems, vol 2062. Springer, Cham. \url{https://doi.org/10.1007/978-3-032-31526-7_18}}}
}

\IfFileExists{xurl.sty}{\usepackage{xurl}}{} 
\IfFileExists{bookmark.sty}{\usepackage{bookmark}}{\usepackage{hyperref}}
\hypersetup{
  hidelinks,
  pdfcreator={LaTeX via pandoc}}
\usepackage{graphicx}
\makeatletter
\def\maxwidth{\ifdim\Gin@nat@width>\linewidth\linewidth\else\Gin@nat@width\fi}
\def\maxheight{\ifdim\Gin@nat@height>\textheight\textheight\else\Gin@nat@height\fi}
\makeatother
\setkeys{Gin}{width=\maxwidth,height=\maxheight,keepaspectratio}
\makeatletter
\def\fps@figure{htbp}
\makeatother
\author{}
\date{}

\DeclareUnicodeCharacter{00B7}{\textperiodcentered}
\DeclareUnicodeCharacter{00D7}{\ensuremath{\times}}
\DeclareUnicodeCharacter{0394}{\ensuremath{\Delta}}
\DeclareUnicodeCharacter{03B1}{\ensuremath{\alpha}}
\DeclareUnicodeCharacter{03B4}{\ensuremath{\delta}}
\DeclareUnicodeCharacter{2265}{\ensuremath{\geq}}
\begin{document}
\thispagestyle{firstpage} 

\fontsize{14}{9}\selectfont\ 
\begin{center} Retrieval-Augmented Generation for Clinical Decision Support in Gait
Analysis: A Proof-of-Concept
\end{center}

\fontsize{11}{9}\selectfont\
\begin{center} Georgios Giarmatzis\textsuperscript{1{[}0000-0003-1387-2134{]}}, Fabian
Horst\textsuperscript{2{[}0000-0002-3299-5896{]}}, Djordje
Slijepcevic\textsuperscript{3{[}0000-0002-2295-7466{]}}, Nikolaos
Aggeloussis\textsuperscript{1{[}0000-0001-5108-7335{]}}, George
Papakostas\textsuperscript{4{[}0000-0001-5545-1499{]}}
\end{center}
\textsuperscript{1} Democritus University of Thrace, Komotini 69100,
Greece

\textsuperscript{2} Johannes Gutenberg-University Mainz, Mainz 55122,
Germany

\textsuperscript{3} St. Pölten University of Applied Sciences, St Pölten
3100, Austria

\textsuperscript{4} Democritus University of Thrace, 65404 Kavala,
Greece\\
ggiarmat@phyed.duth.gr

\textbf{Abstract.} Large language models (LLMs) demonstrate significant
potential for clinical decision support but lack specialized
biomechanical knowledge required for accurate gait analysis
interpretation. We developed and evaluated a Retrieval-Augmented
Generation (RAG) framework to enhance LLM-generated clinical
explanations for ankle fracture rehabilitation using ground reaction
force data. Using the GaitRec dataset (503 subjects: 211 healthy, 292
ankle fracture patients), we implemented a five-stage pipeline
integrating: (1) machine learning classification (91.3\% accuracy); (2)
Layer-wise Relevance Propagation for explainable feature identification;
(3) RAG knowledge base from peer-reviewed biomechanics literature; (4)
GPT-5 explanation generation with and without retrieval augmentation;
and (5) double-blind evaluation by experienced physical therapists
across five clinical criteria. RAG augmentation produced statistically
significant improvements in Clini-cal Benefit (mean difference=0.60,
p=0.047, δ=0.49) and Completeness (mean difference=0.70, p=0.049,
δ=0.49), both demonstrating large effect sizes. Baseline GPT-5 achieved
strong performance across all criteria (\textgreater4.0/5.0). This
proof-of-concept demonstrates that augmenting general-purpose LLMs with
domain-specific knowledge meaningfully enhances actionable,
evidence-based clinical guidance---dimensions most critical for
rehabilitation decision-making. The framework provides a scalable
approach for integrating AI into specialized clinical workflows without
costly model retraining.


\begin{enumerate}
\def\labelenumi{\arabic{enumi}.}
\item
  Literature review
\end{enumerate}

Large language models (LLMs) have been used in healthcare to support
clinical decision and/or diagnosis, alongside medical knowledge
acquisition {[}1--3{]}. Recent research has suggested that LLMs can
provide personalized recommendations for complex medical tasks,
including patient triage and treatment suggestion {[}4{]}. However,
using general LLMs (e.g., GPT-5) in clinical environments can be limited
by factual inaccuracies and lack of domain-specific knowledge that is
needed to explain biomedical analysis outputs {[}5,6{]}. More
specifically, multi-source biomechanical data from clinical gait
analysis require high technical expertise and their interpretation
demand highly specific knowledge {[}7{]}.

Understanding such data is important for assessing the quality of a
patient's walking and discern motion deficits and deviations from normal
{[}9,10{]}. Thus, medical professionals can use this knowledge to plan
effective rehabilitation programs, particularly for patients after an
ankle fracture. However, such biomechanical data are time-dependant and
complex in nature, thus difficult to interpret, like the ground reaction
forces (GRFs) recorded from ground-embedded force plates {[}7,8{]}.
Usage of LLMs can address such challenges, although general-purpose LLMs
might lack the domain-specific knowledge necessary to accurately explain
such data {[}11{]}. This may result in interpretations that are
incomplete, misleading, or potentially harmful to patient care {[}12{]}.

Three primary approaches exist for integrating domain-specific knowledge
into LLMs: (1) fine-tuning general-purpose models with specialized
knowledge, (2) training domain-specific LLMs from scratch, and (3)
Retrieval-Augmented Generation (RAG). The first requires substantial
computational resources {[}13{]}, and the second {[}14{]} demonstrate
strong performance but demand massive domain-specific training datasets
(millions of documents). In contrast, RAG offers a promising, much less
computationally expensive approach to receive a more accurate response
from an LLM, based on an external knowledge base and some prompt
engineering. The RAG workflow operates through three key stages: first,
domain-specific documents (e.g., peer-reviewed research papers on gait
biomechanics) are processed and stored in a structured knowledge base
with semantic embeddings; second, the system retrieves the most relevant
knowledge segments using vector similarity search when presented with a
clinical query; and third, the retrieved context is integrated into the
LLM's prompt to guide response generation {[}4{]}. RAG is particularly
well-suited for specialized clinical domains where knowledge evolves
rapidly, as new research findings can be incorporated into the knowledge
base without retraining the LLM {[}15{]}.

Despite growing interest in applying RAG to clinical domains {[}16{]},
evaluations in specialized areas like clinical gait analysis remain
unexplored. While recent studies demonstrate RAG's potential for
text-based medical tasks {[}17{]}---usually involve retrieving textual
knowledge for textual questions. Clinical gait analysis presents
fundamentally different challenges: interpreting multivariate
time-series biomechanical data and integrating quantitative measurements
with qualitative rehabilitation literature for patient-specific
treatment plans. No prior work has evaluated whether RAG can bridge this
modality gap using expert assessment rather than automated metrics or
patient surveys {[}18{]}. This paper presents the first RAG-enhanced LLM
for clinical biomechanical gait analysis, integrating machine learning
classification, explainable AI, and domain-specific knowledge retrieval,
with double-blind expert evaluation.

\begin{enumerate}
\def\labelenumi{\arabic{enumi}.}
\setcounter{enumi}{1}
\item
  Methodology

  \begin{enumerate}
  \def\labelenumii{\arabic{enumii}.}
  \item
    Dataset
  \end{enumerate}
\end{enumerate}

For our experiments, we employed a subset of the GaitRec dataset
{[}8{]}, which contains ground reaction force (GRF) recordings captured
during overground walking on a level surface along an approximately 10 m
long walkway. The dataset comprises recordings from 503 subjects (211
healthy controls, 292 ankle fracture patients) collected at the AUVA
Rehabilitation Center in Austria between 2007 and 2018 {[}8{]}. GRFs
were captured using two force platforms at 250 Hz during the stance
phase, measuring bilateral forces along three orthogonal axes: vertical
(F\_V), anterior-posterior (F\_AP), and medial-lateral (F\_ML), yielding
six channels per subject. Participants walked at a self-selected speed
up to 10 valid bilateral recordings were obtained, resulting in 2,079
healthy and 2,276 ankle fracture recordings. Demographic variables (sex,
age, height, body mass, affected limb, and footwear) were extracted and
transformed into clinically interpretable formats. The final GRF dataset
was structured into a multidimensional array of dimensions subjects ×
time points × force components.

\begin{enumerate}
\def\labelenumi{\arabic{enumi}.}
\setcounter{enumi}{1}
\item
  Machine Learning for Injury Classification
\end{enumerate}

For the injury classification task (ankle fracture vs. healthy), we
compared different machine learning methods including Multi-layer
Perceptrons (MLPs) with varying depths (2-3 layers) and widths (64-768
units) and Support Vector Machines (SVMs) with different regularization
parameters (C=0.1, 1.0, 10.0). A subject-wise 10-fold cross-validation
scheme was employed, ensuring that all data from a given subject were
contained within a single fold. This setup prevented data leakage across
folds and reduced the likelihood of the model learning subject-specific
gait patterns, thereby promoting the learning of injury-related
features. All model configurations underwent identical training and
evaluation procedures. Among the tested models, the MLP with three
layers and 768 units achieved the highest classification accuracy and
was selected for further analysis.

\begin{enumerate}
\def\labelenumi{\arabic{enumi}.}
\setcounter{enumi}{2}
\item
  Explainable AI Framework
\end{enumerate}

To interpret which biomechanical features the ML models utilized to
distinguish patients with ankle fractures from healthy controls, we
employed Layer-wise Relevance Propagation (LRP), an explainable AI
technique that has been previously applied for gait analysis {[}9,
10{]}. This shows how much each feature adds to the classification
decision. We took relevance scores from the trained model for each
person in the ankle fracture group and we normalized them between 0 and
1. We set a cutoff of \textgreater0.5 to find key biomechanical
patterns, deciding that scores above 0.5 pointed to features that helped
classification and scores below 0.5 showed features that either blocked
or didn't matter for classification. The gait cycle (stance: heel strike
to toe-off) was segmented into four ranges of interest, that is the
loading response (0-25\% of stance), midstance (26-50\%), terminal
stance (51-75\%), and pre-swing (76-100).

\begin{enumerate}
\def\labelenumi{\arabic{enumi}.}
\setcounter{enumi}{3}
\item
  RAG System for Knowledge Integration
\end{enumerate}

The RAG system comprised of three stages: document processing, knowledge
extraction, and retrieval augmentation. Extended search in literature
for papers focusing on post-ankle fracture biomechanics resulted in 24
highly relevant articles.

\emph{Document Processing}: An algorithm was developed that could read
the papers in PDF format. The same algorithm could perform text
extraction and text artifacts like headnotes and footnotes were removed.
Distinct sections were identified (Introduction, Methods, Results,
Discussion) and fed to the next step.

\emph{Knowledge Extraction}: Each section was sent to Claude 3.5 Sonnet
via API to extract six to ten meaningful knowledge chunks. Each chunk
was then classified by type (finding, methodology, limitation,
conclusion) and assigned a confidence score (0.55-0.95).

\emph{Knowledge Storage and Retrieval}: Each chunk was stored in a
database based on the SQLite framework, along with its corresponding
vector produced using the OpenAI's text-embedding-3-small model.

\begin{enumerate}
\def\labelenumi{\arabic{enumi}.}
\setcounter{enumi}{4}
\item
  LLM Framework for Clinical Explanation Generation
\end{enumerate}

The LLM framework included four components: a) the SubjectAnalyzer
component which extracted patient-specific information and which part of
GRFs were used to classify each patient as ankle fracture case, b) the
KnowledgeRetriever component that used the RAG system to populate the
prompt, c) a specialized Prompt Engineering module, and d) the
LLM-Manager component that was used to send the prompt to different LLMs
and receive the responses.

\emph{SubjectAnalyzer:} As mentioned before, this component used a 0.5
cutoff threshold was respected to identify regions of interest in the
relevance scores as outputed from the LRP algorithm, lasting at least
5\% of the gait cycle duration. Those regions that was used to classify
the patient as ankle fracture case, were transformed to meaningful
textual explanations for the LLM.

\emph{KnowledgeRetriever:} This component was mainly used to search for
the most relevant knowledge chunks in the database, based on the user's
query that was also matched with a vector embedding using the OpenAI's
text-embedding-3-small model. The matching was performed combining
cosine similarity (0.6) and confidence scores (0.4).

\emph{Prompt Engineering:} A template was followed during the prompt
construction. Basically, the prompt included the binary classification
result (correct/incorrect) and ground truth label (ankle fracture
patient/healthy control) from the ML algorithm, which enabled the model
to frame explanations appropriately, i.e., acknowledging successful
pattern detection when classification was correct or not. A structured
demographic profile was integrated, including age, biological sex,
height, body mass, affected limb (left/right/bilateral), footwear
condition, and whether an orthopedic insole was worn. LRP-derived
biomechanical features were presented as a prioritized list ordered by
relevance scores. Each feature included four elements: (1) the specific
GRF channel and gait phase (e.g., "vertical force in left leg during
loading response"), (2) peak relevance scores (0-1 scale), (3) temporal
extent (percentage of gait phase affected), and (4) clinical
significance descriptor. For the clinical significance descriptor,
relevance scores were mapped to five interpretable levels: slight
(0.5--0.6), mild (0.6--0.7), moderate (0.7--0.8), strong (0.8--0.9), and
extremely strong (\textgreater0.9), with the latter representing the
primary differentiating factors according to the ML model. When RAG was
activated, domain-specific knowledge was retrieved through semantic
search and prepended to the prompt. Five research queries were
systematically executed: "ankle fracture gait biomechanics ground
reaction forces", "ankle fracture rehabilitation weight bearing
progression", "ankle fracture compensatory movement patterns", "ankle
fracture functional recovery outcomes", and "ground reaction force
analysis gait abnormalities". For each query, the system computed
embedding vectors and performed cosine similarity matching against the
cached knowledge base, retrieving the top-ranked items. Retrieved
knowledge chunks (maximum 20 items across all queries) were formatted as
numbered research context entries, each containing the knowledge type
classification (finding/methodology/limitation), confidence score, and
complete text. We used GPT-5 with and without RAG augmentation, with
each configuration generating two explanations per subject from the
ankle fracture class. The model was instructed to deliver a
comprehensive biomechanical analysis encompassing clinical significance,
underlying mechanisms, and rehabilitation recommendations.

GPT-5 was chosen as the target LLM based on three considerations: First,
it is the current state-of-the-art performance on medical reasoning
studies {[}19{]}. Second, single-model analysis enabled within-model
comparison (with vs. without RAG) while controlling for confounding
variables such as architecture-specific biases. Third, this was a
proof-of-concept study that prioritized evaluation from different
experts rather than comparison between different models.

\begin{enumerate}
\def\labelenumi{\arabic{enumi}.}
\setcounter{enumi}{5}
\item
  Expert Evaluation Framework
\end{enumerate}

Two physical therapists (with over 15 years of clinical experience)
conducted a blinded evaluation of the LLM-generated, patient-specific
rehabilitation protocols using designated Google Forms. Each form
contained demographic data, relevant important areas of the GRF signals
that differentiate each patient, the biomechanical explanation and the
suggested rehabilitation protocol. Five random patients, with respective
protocols generated from GPT-5 with or without RAG (10 evaluations in
total) were assigned in each expert. The evaluation was based on five
criteria using a 1-5 Likert scale: The criteria were a) Accuracy
(measures whether the answer was abides by relevant medical knowledge,
1: very inaccurate; 5: Completely accurate), b) Relevance (checks
whether the answer was directly relevant to the question or it is off
topic, 1: completely irrelevant; 5: Highly relevant), Clarity (measures
whether the answer was easy to understand, 1: Very unclear; 5: Extremely
clear), Clinical benefit (used to assess whether the answer was
significantly helpful in the decision-making process, 1: Not useful; 5:
Extremely Useful), and Completeness (used to assess whether the answer
contains all necessary information to fully answer the question, 1: Very
Incomplete; 5: Completely Complete)

\begin{enumerate}
\def\labelenumi{\arabic{enumi}.}
\setcounter{enumi}{6}
\item
  Data Analysis
\end{enumerate}

All google form results were downloaded and analyzed using custom code.
For every criterion, mean and standard deviation were computed.
Non-parametric Mann-Whitney U test was used to assess statistical
significance (α = 0.05) for each LLM between RAG conditions. Effect
sizes were calculated using Cliff's Delta (δ) equation {[}20{]}. Values
smaller than 0.15 were named negligible, values between 0.15-0.33 were
considered small, between 0.33-0.47 as medium, and larger than 0.47 as
large. Mean differences are reported with 95\% confidence intervals. All
analyses were conducted using Python 3.8.

\begin{enumerate}
\def\labelenumi{\arabic{enumi}.}
\setcounter{enumi}{2}
\item
  Result Analysis
\end{enumerate}

Figure 1 illustrates the LRP analysis for an ankle fracture patient
(Subject 127), displaying GRFs across all six measurement channels
throughout the stance phase. Black traces represent the raw
biomechanical signals, while colored overlays (blue/red) indicate
regions where LRP attributed high relevance scores to the classification
decision, signifying deviation from healthy controls.

\includegraphics[width=4.75479in,height=1.74306in]{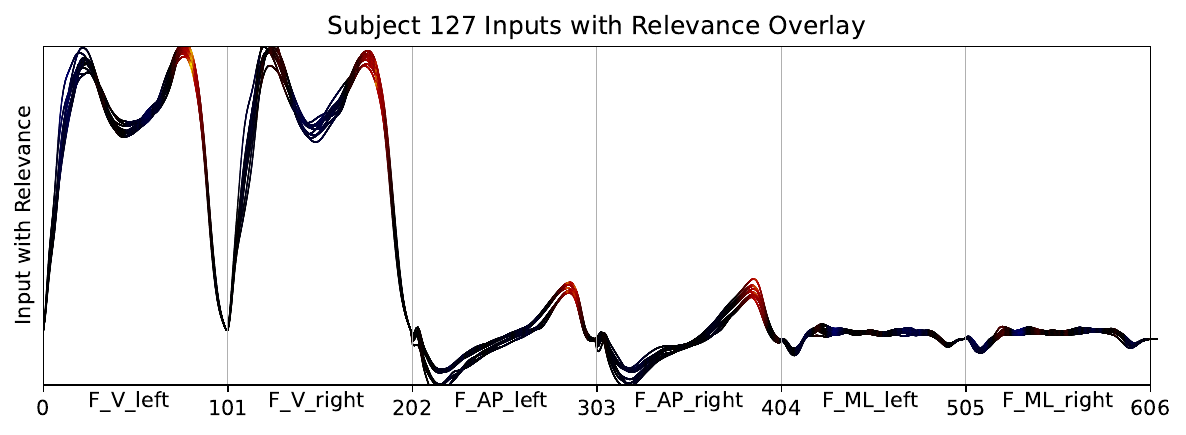}

\textbf{Fig.} \textbf{1.} Ground reaction force patterns with LRP
relevance overlay for Subject 127 (ankle fracture patient). Black traces
display ground reaction forces across vertical (F\_V),
anterior-posterior (F\_AP), and medial-lateral (F\_ML) axes for both
limbs throughout the stance phase. Superimposed color gradients
represent LRP relevance values, with red tones indicating features
contributing positively to the ankle-fracture prediction and blue tones
indicating features opposing it.

\includegraphics[width=4.79861in,height=2.77499in]{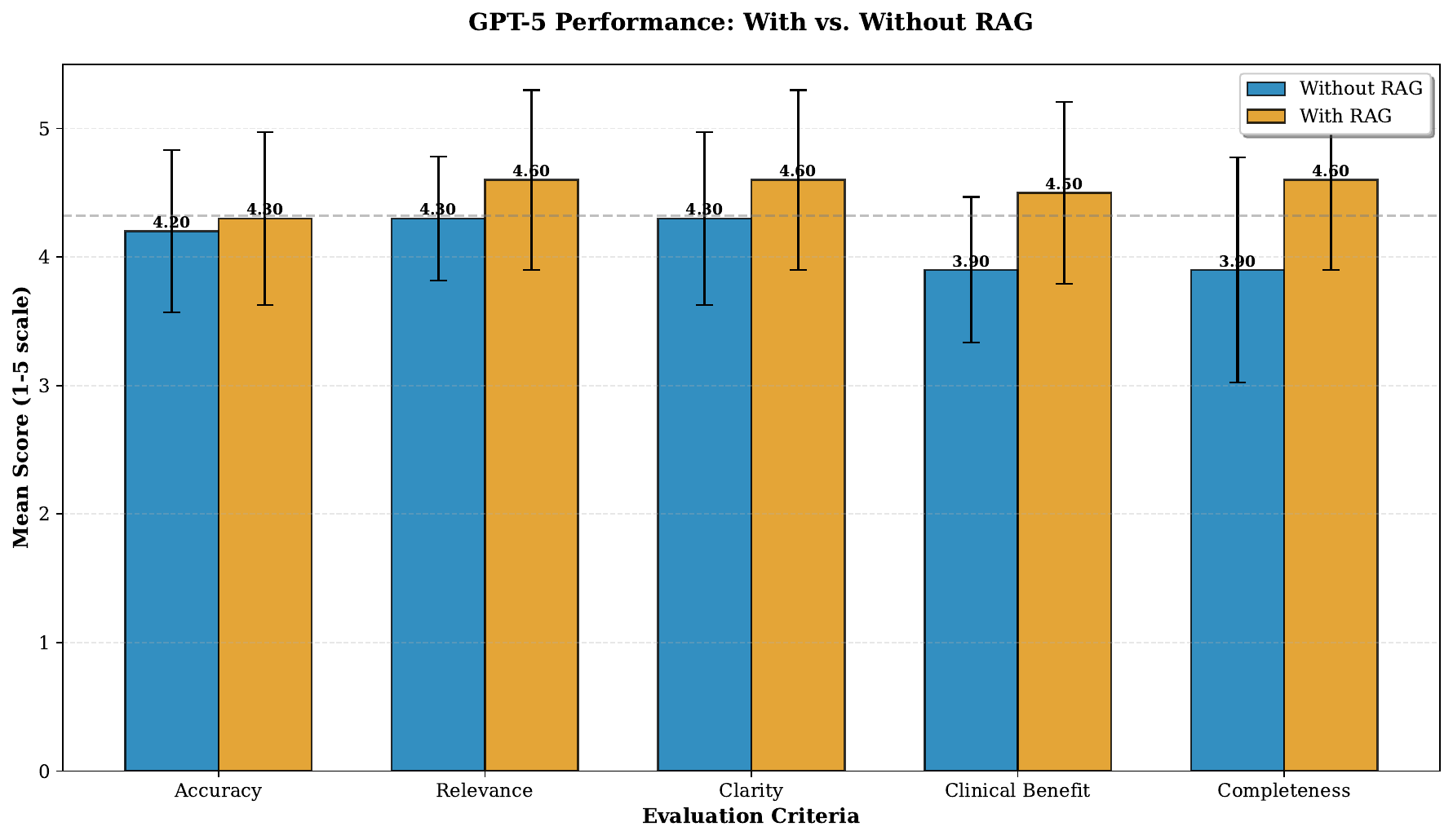}

\textbf{Fig.} \textbf{2.} Expert evaluation ratings comparing GPT-5
performance with and without RAG augmentation across five criteria.
Green bars represent RAG-enabled explanations; red bars represent
baseline explanations. Error bars indicate standard deviation (n=20 per
condition).

Expert evaluations (n=20 per condition) revealed consistently high
ratings for GPT-5-generated explanations across all criteria, with mean
scores ranging from 4.20 to 4.60 on the 5-point scale (Figure 2). RAG
augmentation produced numerical improvements across all five dimensions,
with two criteria achieving statistical significance.Non-significant
improvements were observed for Accuracy (Δ=0.10, p=0.734, negligible
effect), Relevance (Δ=0.30, p=0.172, medium effect), and Clarity
(Δ=0.30, p=0.269, small effect), where baseline GPT-5 already
demonstrated strong performance (M≥4.20). Statistically significant
improvements emerged for Clinical Benefit (Δ=0.60, p=0.047, δ=0.49) and
Completeness (Δ=0.70, p=0.049, δ=0.49), both showing large effect sizes.
These results indicate that RAG augmentation meaningfully enhances the
two dimensions most critical for clinical actionability---providing
evidence-based guidance and comprehensive coverage of rehabilitation
considerations---while maintaining the accuracy and clarity strengths of
the baseline model. Mean values are shown in Figure 2.

\textbf{Qualitative Example of RAG Enhancement.} Subject 1858
(23-year-old female, right ankle fracture) demonstrated bilateral
pre-swing force abnormalities.

Baseline GPT-5 provided general recommendations:

\begin{itemize}
\item
  \begin{quote}
  "Manual mobilizations to regain dorsiflexion"
  \end{quote}
\item
  \begin{quote}
  "Seated and standing heel-raises, progressing from bilateral to
  unilateral"
  \end{quote}
\item
  \begin{quote}
  "Balance drills on stable to progressively unstable surfaces"
  \end{quote}
\item
  \begin{quote}
  "Reassess GRF symmetry and ankle kinematics every 2--4 weeks"
  \end{quote}
\end{itemize}

RAG-enhanced GPT-5 integrated evidence-based protocols with specific
timelines:

\begin{itemize}
\item
  \begin{quote}
  Phase 1 (Weeks 1--4 post-op): "Early protected weight-bearing as
  tolerated, progressing from shoe/support to normal shoe by week 2--3"
  \end{quote}
\item
  \begin{quote}
  Phase 2 (Weeks 4--8): "Treadmill or overground gait retraining with
  real-time feedback (mirror or force-plate visual) to restore symmetry
  in vertical and A-P force symmetry in pre-swing"
  \end{quote}
\item
  \begin{quote}
  Phase 3 (Weeks 8--12+): "Plyometric drills... once GRF symmetry
  \textgreater90\%"
  \end{quote}
\item
  \begin{quote}
  Ongoing goals: "Achieve \textless10\% side-to-side difference in peak
  vertical and A-P forces during pre-swing"
  \end{quote}
\end{itemize}

\begin{enumerate}
\def\labelenumi{\arabic{enumi}.}
\setcounter{enumi}{3}
\item
  Conclusion
\end{enumerate}

This proof-of-concept study evaluated the effectiveness of RAG in
enhancing LLM performance for generating clinical explanations of
biomechanical gait analysis data in ankle fracture patients. Expert
evaluations revealed that GPT-5 generated high-quality clinical
explanations across all evaluation criteria, with mean scores ranging
from 4.20 to 4.60 on a 5-point Likert scale, indicating that even
without domain-specific augmentation, modern LLMs demonstrate
substantial baseline competency in interpreting complex biomechanical
data. This finding agrees with recent findings showing that general LLMs
can achieve physician-level accuracy on complex medical reasoning tasks
when provided with well-structured clinical data {[}21{]}. However, RAG
augmentation produced statistically significant improvements in two
clinically critical dimensions: Clinical Benefit (mean difference = 0.60
points, 95\% CI: 0.04-1.16, p=0.047, δ=-0.49) and Completeness (mean
difference = 0.70 points, 95\% CI: 0.01-1.39, p=0.049, δ=-0.49), both
demonstrating medium effect sizes. The evidence gathered indicates that
enhancing broad-scope language models by incorporating domain-relevant
facts extracted from scholarly publications substantially boosts their
capacity to formulate rehabilitation recommendations that are both
practical for clinical use and thorough.

The improvements observed in Clinical Benefit and Completeness need to
be further discussed. Clinical Benefit, defined as the potential to
improve patient outcomes on the basis of the rehabilitation
recommendations, represents a pragmatic measure of our system's utility
in real-world clinical settings.. The 0.60-point improvement in this
criterion suggests that RAG-augmented, AI generated rehabilitation
protocols were specific enough that physical therapists could directly
implement. Similarly, the 0.70-point improvement in Completeness
indicates that the explanation provided a broader spectrum of clinically
relevant factors like compensatory movement patterns, that might
otherwise be overlooked when relying solely on the LLM's pre-trained
knowledge. These improvements are especially relevant due to the
variability of ankle fracture cases and the current lack of
standardized, evidence-based rehabilitation protocols in current
clinical practice.

This study addresses a critical gap in literature by providing one of
the first rigorous, expert-based evaluations of RAG systems in a highly
specialized clinical domain, i.e., biomechanical gait analysis. While
recent studies have demonstrated the potential of RAG to enhance LLM
performance in general medical knowledge retrieval {[}22{]}, diagnostic
assistance {[}23{]}, and treatment recommendation tasks, these
applications have predominantly focused on text-based clinical scenarios
where domain knowledge can be straightforwardly retrieved from medical
literature or clinical guidelines. The present work extends RAG
methodology into the complex territory of multivariate time-series
biomechanical data interpretation, where prompts must integrate
quantitative force measurements, temporal gait phase dynamics, and
patient-specific demographic factors with domain-specific retrieved
knowledge. Our findings support the work by Chen et al. {[}4{]}
regarding RAG's potential for treatment decision-making in
musculoskeletal conditions. Also, we provide empirical evidence that
such systems can generate explanations that are clinically valuable by
experienced practitioners. Moreover, by incorporating LRP as an
explainability layer, this framework addresses the critical "black box"
problem that has historically limited clinical adoption of deep learning
systems {[}24{]}. The range of effect sizes observed in this study are
particularly encouraging given the small sample size and the
already-high baseline performance of GPT-5, suggesting that larger
validation studies may reveal more substantial benefits when RAG
addresses specific knowledge deficiencies in the base model's training
data.

The current study has some limitations. First, the small sample size
(n=20 evaluations per criterion) limits statistical power, particularly
for detecting small-to-medium effect sizes. This limitation likely
explains why metrics such as Relevance, despite demonstrating a medium
effect size (δ=0.33), did not achieve statistical significance. Larger
validation studies are needed to establish whether these numerical
improvements represent genuine effects or sampling variability. Second,
while the methods describe a single-model framework using GPT-5, a
systematic comparison of different LLM architectures' responses to the
same questions, with or without RAG, is necessary. Third, the small
number of experts limits the generalizability of quality assessments
despite the blinded evaluation design, even though both of them had
extensive clinical experience (\textgreater15 years). Finally, the
limited number of peer-reviewed publications on ankle fracture gait
biomechanics (only 24), may not have been sufficient to capture the full
scope of relevant literature.

This proof-of-concept study provides novel evidence that RAG can enhance
LLM performance for generating clinical explanations in ankle fracture
biomechanical gait analysis, with statistically significant improvements
estimated in Clinical Benefit and Completeness. By augmenting the prompt
sent to GPT-5 with domain-specific knowledge, the current study has
demonstrated a viable pathway for augmenting clinical decision support
without costly model (re)training {[}25{]}. This way the fundamental
challenge of LLM knowledge retrieval in rapidly evolving specialized
medical domains can be addressed. RAG systems could be used in various
clinical scenarions where they can function as real-time decision
support tools during patient consultations, providing clinicians with
AI-generated rehabilitation protocols. Future work should focus on: (1)
inclusion of more experts with different but relevant backgrounds to
musculoskeletal rehabilitation; (2) comparison of different LLM
architectures to verify if RAG enhancements are model-agnostic; and (3)
prospective clinical trials that will use similar frameworks to
rehabilitate patients and assess their outcome. Establishing
applicability to further conditions affecting the musculoskeletal
system, such as anterior cruciate ligament repair or complete knee joint
replacement, alongside integration with digital patient recors, is
crucial. This step is necessary to validate the framework's
applicability and to facilitate its real-world clinical implementation.
Nonetheless, our work's findings showcase the strength of RAG-enabled
prompt enhancement to acquire improved rehabilitation protocols and
assist clinicians in interpreting complex biomechanical data.

Acknowledgments

This work was supported by the MPhil program ``Advanced Technologies in
Informatics and Computers'', which was hosted by the Department of
Informatics, Democritus University of Thrace, Kavala, Greece. This work
was partly funded by the Austrian Research Promotion Agency of Lower
Austria (GFF, Gesellschaft für Forschungsförderung NÖ) within the
deepForce project (GLF21-1-009).

\textbf{References}

1. Vrdoljak, J.; Boban, Z.; Vilović, M.; et al. A Review of Large
Language Models in Medical Education, Clinical Decision Support, and
Healthcare Administration. \emph{Healthc. 2025, Vol. 13, Page 603}
\textbf{2025}, \emph{13}, 603.

2. Zhang, K.; Meng, X.; Yan, X.; et al.
Revolutionizing Health Care: The Transformative Impact of Large Language
Models in Medicine. \emph{J. Med. Internet Res.} 2025, \emph{27},
e59069, doi:10.2196/59069.

3. Gaber, F.; Shaik, M.; Allega, F.; et al.Evaluating large language model workflows in clinical decision support for triage and referral and diagnosis. \emph{npj Digit. Med.} \textbf{2025}, \emph{8}, 1--14.

4. Chen, R.; Zhang, S.; Zheng, Y.; et al. Enhancing treatment decision-making for low back pain: a novel framework integrating large language models with retrieval-augmented generation technology. \emph{Front. Med.} \textbf{2025}, \emph{12}, 1599241.

5. Eke, C.I.; Shuib, L. The role of explainability and transparency in fostering trust in AI healthcare systems: a systematic literature review, open issues and potential solutions. \emph{Neural Comput. Appl. 2024 374} \textbf{2024}, \emph{37}, 1999--2034.

6. Amann, J.; Blasimme, A.; Vayena, E.; et al. Explainability for artificial intelligence in healthcare: a multidisciplinary perspective. \emph{BMC Med. Inform. Decis. Mak.} \textbf{2020}, \emph{20}, 1--9.

7. Rebelo, A.; Martinho, D. V.;
Valente-dos-Santos, J.; et al. From data to action: a scoping review of
wearable technologies and biomechanical assessments informing injury
prevention strategies in sport. \emph{BMC Sports Sci. Med. Rehabil.}
\textbf{2023}, \emph{15}, 1--14.

8. Horsak, B.; Slijepcevic, D.;
Raberger, A.M.; et al. GaitRec, a large-scale ground reaction force
dataset of healthy and impaired gait. \emph{Sci. Data 2020 71}
\textbf{2020}, \emph{7}, 1--8.

9. Horst, F.; Lapuschkin, S.; Samek, W.;
et al. Explaining the unique nature of individual gait patterns with
deep learning. \textbf{2019}, 1--13.

10. Slijepcevic, D.; Horst, F.;
Lapuschkin, S.; et al. Explaining Machine Learning Models for Clinical
Gait Analysis. \emph{ACM Trans. Comput. Healthc.} \textbf{2021},
\emph{3}, 14.

11. Zheng, W.; Arakawa, Y. A Review of Wearables-Based
Activities of Daily Living Recognition With LLMs: Overview, Progress,
and Trends. \emph{IEEE Access} \textbf{2025}, \emph{13},
143813--143830.

12. Bonnechère, B. Unlocking the Black Box? A
Comprehensive Exploration of Large Language Models in Rehabilitation.
\emph{Am. J. Phys. Med. Rehabil.} \textbf{2024}, \emph{103},
532--537.

13. Gu, Y.; Tinn, R.; Cheng, H.; et al. Domain-Specific
Language Model Pretraining for Biomedical Natural Language Processing.
\emph{ACM Trans. Comput. Healthc.} \textbf{2021}, \emph{3}, 24.

14. Tu,
T.; Azizi, S.; Driess, D.; et al. Towards Generalist Biomedical AI.
\emph{NEJM AI} \textbf{2024}, \emph{1}.

15. Chan, B.J.; Chen, C.-T.;
Cheng, J.-H.; et al. Don't Do RAG: When Cache-Augmented Generation is
All You Need for Knowledge Tasks. \textbf{2025}, 893--897.

16. Zhao, S.;
Yang, Y.; Wang, Z.; et al. Retrieval Augmented Generation (RAG) and
Beyond: A Comprehensive Survey on How to Make your LLMs use External
Data More Wisely. \textbf{2024}.

17. Van Veen, D.; Van Uden, C.;
Blankemeier, L.; et al. Adapted large language models can outperform
medical experts in clinical text summarization. \emph{Nat. Med. 2024
304} \textbf{2024}, \emph{30}, 1134--1142.

18. Ayers, J.W.; Poliak, A.;
Dredze, M.; et al. Comparing Physician and Artificial Intelligence
Chatbot Responses to Patient Questions Posted to a Public Social Media
Forum. \emph{JAMA Intern. Med.} \textbf{2023}, \emph{183}, 589--596.

19.
Wang, S.; Hu, M.; Li, Q.; et al. Capabilities of GPT-5 on Multimodal
Medical Reasoning. \textbf{2025}.

20. Meissel, K.; Yao, E.S. Using
Cliff's Delta as a Non-Parametric Effect Size Measure: An Accessible Web
App and R Tutorial. \emph{Pract. Assessment, Res. Eval.} \textbf{2024},
\emph{29}, 1--12.

21. Singhal, K.; Azizi, S.; Tu, T.; et al. Large
language models encode clinical knowledge. \emph{Nat. 2023 6207972}
\textbf{2023}, \emph{620}, 172--180.

22. Shi, Y.; Xu, S.; Yang, T.; et
al. MKRAG: Medical Knowledge Retrieval Augmented Generation for Medical
Question Answering. \emph{AMIA Annu. Symp. Proc.} \textbf{2025},
\emph{2024}, 1011.

23. Pinna, S.; Massa, S.M.; Riboni, D. AI-Driven
Differential Diagnosis: Leveraging RAG and LLMs in Intelligent
Healthcare Systems. \emph{2025 21st Int. Conf. Intell. Environ. IE 2025}
\textbf{2025}.

24. Amann, J.; Blasimme, A.; Vayena, E.; et al.
Explainability for artificial intelligence in healthcare: a
multidisciplinary perspective. \emph{BMC Med. Inform. Decis. Mak.}
\textbf{2020}, \emph{20}.

25. Bommasani, R.; Hudson, D.A.; Adeli, E.; et
al. On the Opportunities and Risks of Foundation Models. \textbf{2021}.

\end{document}